\documentclass[runningheads]{llncs}

\usepackage{eccv}

\usepackage{eccvabbrv}
\usepackage{graphicx}
\usepackage{booktabs}
\usepackage{multirow}
\usepackage{tikz}
\usetikzlibrary{arrows.meta,backgrounds,fit,positioning}
\usepackage{algorithm}
\usepackage{algpseudocode}
\usepackage[accsupp]{axessibility}

\usepackage[pagebackref,breaklinks,colorlinks,citecolor=eccvblue]{hyperref}

\begin{document}

\title{From Visual Widgets to UI Code: Efficient Tool-Grounded Generation}
\titlerunning{Efficient Tool-Grounded Widget-to-Code Generation}

\author{Houston H. Zhang$^{1}$, ~ 
Tao Zhang$^{2}$, ~
Li Gu$^{3}$, ~
Linfeng Ye$^{2}$, ~
Yuanhao Yu$^{1}$, ~
Xinxin Zuo$^{3}$, ~
Yang Wang$^{3}$, ~
Zhixiang Chi$^{2\dag}$
}
\authorrunning{Zhang et al.}

\institute{${^1}$ McMaster University ~
${^2}$ University of Toronto ~
${^3}$ Concordia University
}
\begingroup
\renewcommand{\thefootnote}{\(\dagger\)}
\footnotetext{Project Lead}
\endgroup
\maketitle

\begin{abstract}
Existing screenshot-to-code systems face a trade-off between flexibility and controllability.
Direct multimodal generation can hallucinate visible details, whereas structured pipelines reduce such errors through component-wise decomposition, predefined templates, and customized intermediate representations.
These structures, however, introduce additional generative orchestration and restrict outputs to designs covered by the representation.
We investigate whether selective tool grounding can improve the fidelity--efficiency trade-off of direct widget-to-code generation.
We introduce \textbf{WidgetGen}, a lightweight tool-grounded framework that extracts observable text and color evidence, performs high-level layout and optional chart reasoning, and directly generates executable JavaScript XML (\emph{JSX}).
This design reduces reliance on component-wise generation while avoiding a fixed UI schema.
Across six multimodal models and \(1{,}000\) held-out widgets, WidgetGen outperforms direct prompting and the structured Widget2Code pipeline on most visual reconstruction metrics, with consistent gains in area, legibility, and style.
Finally, reconstruction-derived image-code pairs improve six Qwen-family open-weight models across every reported metric through supervised fine-tuning.
These results establish WidgetGen as a strong lightweight baseline and show that selective evidence grounding offers an effective alternative to extensive representation constraints.
\keywords{Widget-to-code generation \and Multimodal large language models \and Tool-grounded generation}

\end{abstract}

% !TEX root = ../main.tex

\section{Introduction}

Visual widgets, such as status cards, metric tiles, and compact charts, are ubiquitous across mobile systems, web dashboards, and applications.
During design handoff, these elements may be available only as screenshots, whereas deployment requires executable UI code.
Following the Widget2Code task formalized by \cite{widgetfactory}, we study how to generate code whose rendered output reproduces a given widget screenshot.
Unlike page-scale interfaces, widgets combine text, dense iconography, data visualizations, and distinctive styling within a small canvas, where a single incorrect element can dominate the reconstruction.
They also typically lack accessible source code, layout metadata, structural annotations, public image--code pairs, and even clean isolated images \cite{widgetfactory,chi2021test}.

Recent multimodal large language models (MLLMs), including GPT-4o, Gemini 3 Pro, and Claude Opus 4.5, accept both image and text inputs \cite{gpt4o,gemini3pro,claudeopus45}.
Screenshot-to-code research has consequently progressed from learned generation on synthetic graphical interfaces \cite{pix2code}, to large-scale paired webpage screenshots and HTML code \cite{websight}, and to frontier MLLMs evaluated on real-world webpages \cite{design2code}.
Direct generation preserves the expressivity of general-purpose UI code, but entangles visual perception, spatial reasoning, and program synthesis in a single response.
Errors in text, color, iconography, layout, or chart content therefore propagate directly to the rendered output \cite{widgetfactory}.

Widget2Code demonstrates that task-specific structure can substantially improve controllability.
It decomposes widgets into atomic components, retrieves icons, instantiates reusable visualization templates, represents the result in WidgetDSL, compiles the intermediate representation deterministically, and adaptively renders the output \cite{widgetfactory}.
This is a strong solution to the fidelity problem because control is encoded in both the representation and execution pipeline.
It also introduces a complementary trade-off: fine-grained generation requires additional orchestration, while supporting a visual pattern outside the predefined components, templates, or intermediate representation can require extending the infrastructure.
This trade-off raises a natural question: \emph{can selective tool grounding improve the fidelity--efficiency trade-off of direct widget-to-code generation?}

Our key observation is that not every reconstruction subproblem requires generation.
Visible text and dominant colors can be extracted from the screenshot, widget dimensions are known, and renderability can be checked through execution.
Providing this evidence explicitly allows the MLLM to focus its capacity on the ambiguous decisions: interpreting layout, reconstructing charts, and synthesizing code.
We hypothesize that selective evidence grounding can recover much of the controllability of a structured pipeline while reducing task-specific representation machinery.

Based on this principle, we introduce \textbf{WidgetGen}, a lightweight tool-grounded framework.
WidgetGen extracts text and color evidence through optical character recognition (\emph{OCR}) and palette analysis, performs high-level layout and optional chart reasoning, directly generates executable JavaScript XML (\emph{JSX}), and renders the result in a browser.
Unlike Widget2Code, it does not require explicit component-wise generation, a predefined template library, a customized intermediate UI representation, or a task-specific compiler.
The framework therefore shifts control from constraining the output representation toward grounding the generator with observable input evidence, while retaining the flexibility of general-purpose UI code.

Our contributions are threefold.
First, we formulate selective evidence grounding as a design principle that separates observable visual facts from ambiguous generative decisions.
Second, we instantiate this principle in WidgetGen and establish a lightweight alternative to component-wise, schema-constrained widget generation.
Third, we provide a systematic evaluation across six MLLMs on 1,000 held-out widgets against both single-shot prompting and the structured Widget2Code pipeline.
WidgetGen obtains the strongest results on most reconstruction metrics, with consistent improvements in Area, legibility, and style.
As a secondary result, we show that reconstruction-derived image--code pairs provide effective supervision for fine-tuning open-weight Qwen models.

% !TEX root = ../main.tex
\section{Related Work}

\label{sec:rw}

\subsection{Multimodal Large Language Models}

Many open-weight multimodal large language models (MLLMs) connect a visual encoder to a large language model through a learned cross-modal interface.
BLIP-2 bridges frozen vision and language backbones with a lightweight Querying Transformer, reducing the trainable parameters required for vision--language pre-training \cite{blip2}, while LLaVA introduces visual instruction tuning for a multimodal model that connects a vision encoder and an LLM for general-purpose visual and language understanding \cite{llava}.
More recent architectures further strengthen multimodal perception and reasoning: Qwen3-VL uses DeepStack to integrate multi-level ViT features for tighter vision--language alignment and enhanced interleaved-MRoPE for stronger spatial--temporal modeling \cite{qwen3vl}, whereas GPT-4o is trained end-to-end across text, vision, and audio \cite{gpt4o}.
These advances make MLLMs capable foundations for visual interface understanding and code synthesis, but general-purpose multimodal competence does not by itself ensure high-fidelity reconstruction.
Widget-to-code generation requires simultaneously recovering dense text, colors, layout, icons, and charts, and expressing these elements as executable code.
Our method therefore uses MLLMs as reasoning and generation backbones, with OCR and palette measurements providing explicit visual evidence for interface reconstruction.

\subsection{UI-to-Code Generation}

Progress has also been supported by broader data and evaluation protocols.
Web2Code provides a large-scale webpage-to-code dataset for instruction tuning and an evaluation framework for webpage understanding and HTML code translation \cite{web2code}.
WebCode2M provides 2.56 million real-world webpage instances with design images, source code, and layout information \cite{webcode2m}.
Design2Code evaluates reconstruction of 484 real-world webpages with render-based metrics and human judgments \cite{design2code}; WebUIBench evaluates WebUI perception, HTML programming, WebUI--HTML understanding, and WebUI-to-code generation using 21K high-quality question--answer pairs derived from more than 700 real-world websites \cite{webuibench}; and DesignBench extends evaluation across React, Vue, Angular, and HTML/CSS to generation, editing, and repair \cite{designbench}.
ChartMimic complements page-oriented benchmarks with 4,800 chart--instruction--code triplets and metrics at both code and rendered-chart levels \cite{chartmimic}.
Together, these settings cover webpages, framework-specific development, and individual charts; compact widgets instead place text, icons, styling, layout, and charts within the same constrained canvas.

Against this empirical background, UI-to-code methods differ in where they impose task-specific structure: a model may generate code directly, or a system may introduce explicit spatial decomposition, modular stages, or a constrained intermediate representation.

\textbf{Direct image-to-code generation.}
Direct approaches take a UI image as input and produce executable implementation code without an explicit intermediate representation or an externally orchestrated decomposition stage at inference time.
WebSight scales this paradigm with two million synthetic screenshot--HTML pairs for training image-to-HTML models \cite{websight}.
WAFFLE strengthens direct HTML generation through structure-aware attention and contrastive image--code alignment \cite{waffle}, while Flame constructs React-oriented training data through component extraction and multiple synthesis strategies, then trains a vision-language model to generate React code from UI images \cite{flame}.
Although these methods may encode structural priors during training, their inference path remains a direct mapping from the input image to implementation code.
Learning this direct mapping remains challenging because models must recover salient structure and fine-grained visual details from pixels while simultaneously producing executable code.

\textbf{Spatial and hierarchical decomposition.}
Structured systems make spatial partitioning or interface hierarchy explicit before producing the complete program.
DCGen partitions screenshots into manageable segments and reassembles segment-level descriptions into complete UI code \cite{dcgen}.
LaTCoder generates code for image blocks with Layout-as-Thought prompting and dynamically selects an assembly strategy \cite{latcoder}.
UICopilot decouples coarse HTML hierarchy prediction from fine-grained code generation \cite{uicopilot}, while LayoutCoder constructs element relations and a UI layout tree to guide code generation and fusion \cite{ui2code_layout}.
Although decomposition reduces whole-interface generation complexity, it introduces intermediate dependencies on segmentation, hierarchy prediction, and assembly, through which early structural errors can propagate to the final implementation.

\textbf{Feedback-driven refinement.}
Another line of work improves UI reconstruction by comparing rendered outputs with target designs and iteratively correcting the generated code.
UIOrchestra employs a difference-analysis agent to identify discrepancies between rendered pages and input designs and guide subsequent revision \cite{uiorchestra}.
DesignCoder recovers component hierarchies from metadata-rich design mockups and applies a vision-guided render-and-compare loop for self-correction \cite{designcoder}, while UI2Code$^{N}$ formulates UI-to-code generation as an interactive process of execution, visual inspection, and iterative refinement \cite{ui2coden}.
These methods place control after initial generation by using visual execution feedback to diagnose and correct reconstruction errors, analogous to test-time adaptation~\cite{ye2026asmil,wu2024test,chi2025learning}.
However, repeated refinement incurs additional inference and rendering cost, and translating visual discrepancies into precise code edits remains challenging.

\textbf{Constrained intermediate representations.}
Other methods express generated interfaces through task-specific structures that can be rendered or compiled systematically.
UI-UG jointly trains UI understanding and generation in a unified MLLM and uses an LLM-friendly UI DSL in its generation workflow \cite{uiug}.
Widget2Code combines atomic component analysis, icon retrieval, reusable visualization modules, WidgetDSL, deterministic compilation, and adaptive rendering \cite{widgetfactory}.
These representations provide an explicit contract between visual interpretation and implementation.
Their controllability, however, comes with representation and maintenance costs: structures outside the predefined schema must be approximated or require coordinated extensions to the vocabulary and rendering infrastructure.

WidgetGen explores an alternative point in this design space: it grounds high-level layout and chart reasoning with extracted text and color evidence before the MLLM directly synthesizes executable JSX.
This design combines visual evidence with direct code generation without relying on a customized intermediate UI representation.

% !TEX root = ../main.tex

%%%%%%%%%%%%%%%%%
\section{Methodology}

\subsection{Task Definition}
\label{sec:task}
Given a target widget screenshot \(I\) of size \(W \times H\), the task is to generate executable UI code \(P\) whose browser rendering \(\hat{I} = R(P, W, H)\) matches the visible content and appearance of \(I\).
WidgetGen additionally uses OCR evidence \(\mathcal{T}\), containing recognized text and bounding boxes, and palette evidence \(\mathcal{C}\), containing dominant colors and their pixel coverage; both are extracted from \(I\).

\subsection{WidgetGen Framework}
\label{sec:overview}

\begin{figure}[t]
\centering
\includegraphics[width=\textwidth]{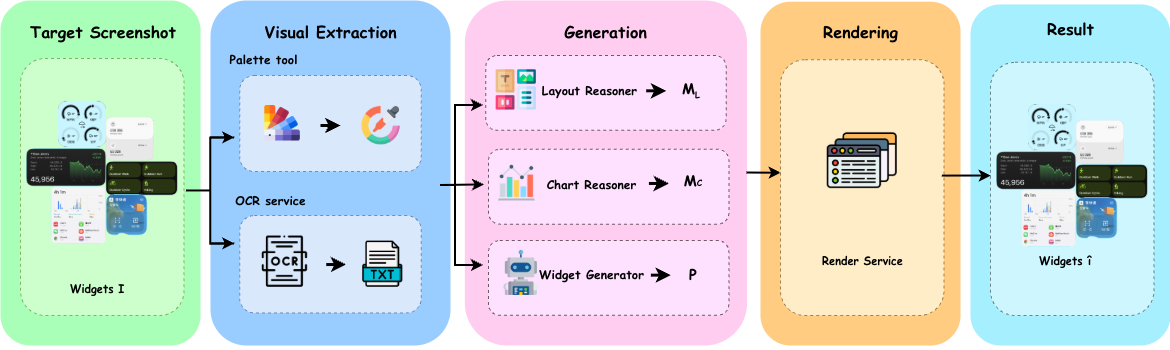}
\caption{Overview of WidgetGen. Extracted text and color evidence ground high-level layout and chart reasoning before direct JSX synthesis and browser rendering.}
\label{fig:arch}
\end{figure}

WidgetGen separates directly observable visual facts from decisions that require generative reasoning.
As shown in \Cref{fig:arch}, text and color evidence are first extracted from the screenshot.
An MLLM then reasons about the whole-widget layout, optionally reconstructs chart content, and directly synthesizes executable React JSX.
The resulting program is rendered once at the target dimensions.

This design places explicit control on the evidence supplied to the generator rather than on a task-specific output representation.
WidgetGen does not perform component-wise generation and does not require a predefined component inventory, a customized UI schema, or a task-specific compiler.
The extracted evidence is shared across layout reasoning, chart reconstruction, and code synthesis, preserving a compact path from the screenshot to executable output.

\subsection{Evidence Extraction}
\label{sec:tools}

The evidence layer targets visual facts that can be measured before code generation.
For a screenshot \(I\), OCR and palette analysis produce
\begin{equation}
\label{eq:evidence}
\begin{aligned}
\mathcal{T} &= \Phi_{\mathtt{ocr}}(I)
             = \{(t_j,b_j)\}_{j=1}^{n_T}, \\
\mathcal{C} &= \Phi_{\mathtt{pal}}(I)
             = \{(c_k,r_k)\}_{k=1}^{n_C}, \qquad n_C \leq 8,
\end{aligned}
\end{equation}
where \(t_j\) and \(b_j\) are a recognized text string and its bounding box, while \(c_k\) and \(r_k\) are a hexadecimal color value and its approximate pixel coverage.
The framework retains the observed dimensions \((W,H)\) for terminal rendering.

\paragraph{Text evidence.}
The text extractor \(\Phi_{\mathtt{ocr}}\) localizes visible strings and represents each observation with its recognized content \(t_j\) and bounding box \(b_j\).
This evidence supports both lexical fidelity and spatial placement, reducing ambiguity when small or densely arranged labels are difficult to recover from visual features alone.

\paragraph{Color evidence.}
The palette extractor \(\Phi_{\mathtt{pal}}\) summarizes the screenshot using dominant colors \(c_k\) and their approximate pixel coverage \(r_k\).
The resulting evidence captures the principal background, foreground, and accent colors while suppressing minor pixel-level variation, providing explicit chromatic cues for code generation.

\begin{figure}[t]
\centering
\includegraphics[width=0.82\textwidth]{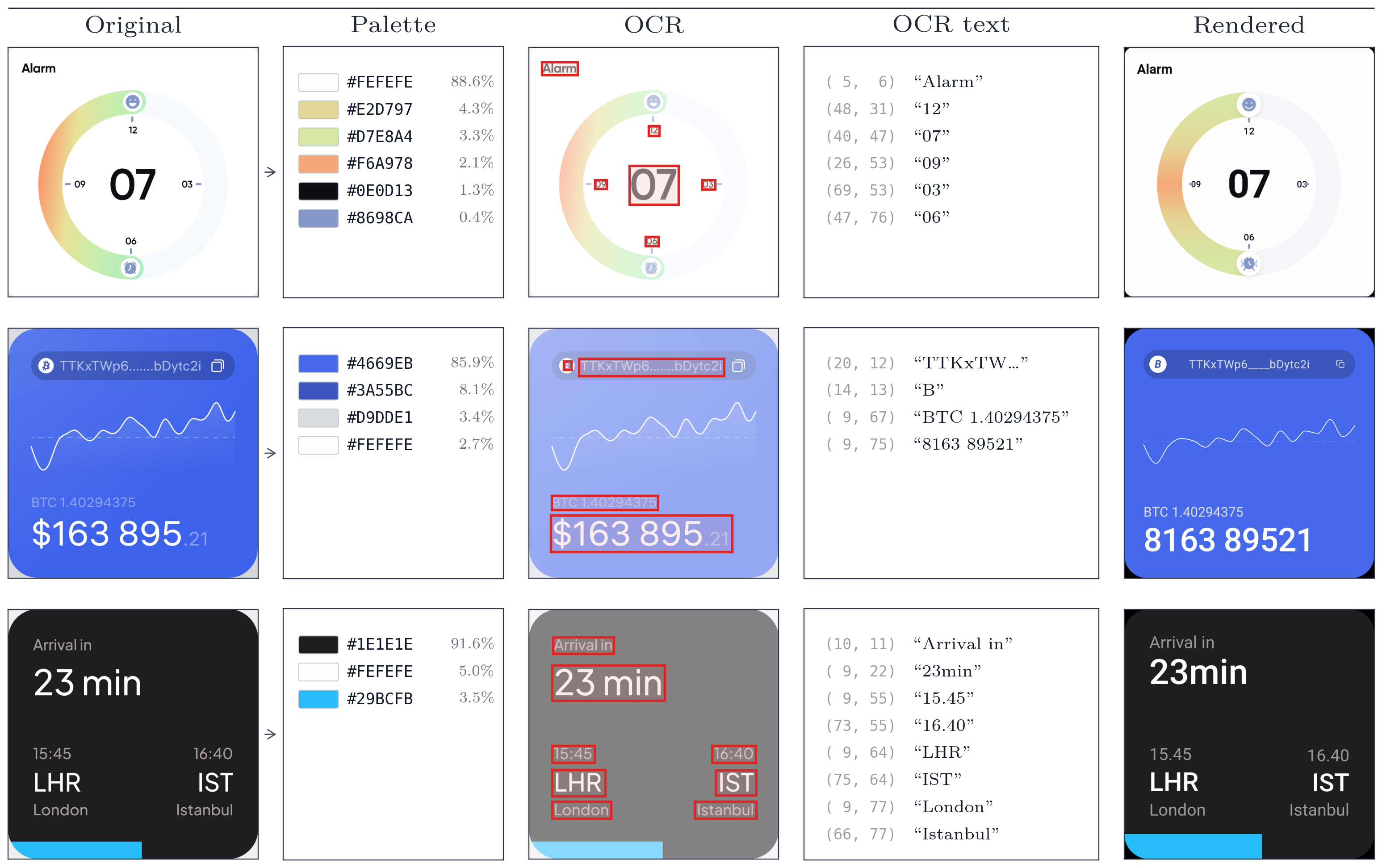}
\caption{Examples of evidence extracted from target widgets. OCR provides recognized strings and locations, while palette analysis provides dominant colors and their coverage.}
\label{fig:tool_results}
\end{figure}

\Cref{fig:tool_results} visualizes the resulting text and palette evidence.
These outputs are serialized into the prompts to ground the subsequent reasoning and code-generation stages.

\subsection{Grounded Code Generation}
\label{sec:gen}

WidgetGen conditions each generative decision on the original screenshot and the extracted evidence.
\Cref{alg:widgetgen} summarizes the inference procedure.

\begin{algorithm}[t]
    \caption{WidgetGen inference for one widget.}
    \label{alg:widgetgen}
    \begin{algorithmic}[1]
        \State \textbf{Input:} target screenshot \(I\) of size \(W \times H\).
        \State \textbf{Output:} JSX program \(P\) and terminal rendering \(\hat{I}\).
        \State In parallel: \(\mathcal{T} \gets \Phi_{\mathtt{ocr}}(I)\), \(\mathcal{C} \gets \Phi_{\mathtt{pal}}(I)\).
        \State \(M_L \gets f_L(I,\mathcal{T},\mathcal{C})\). \Comment{whole-widget layout}
        \If{\(\mathbb{I}_{\mathtt{chart}}(M_L)=1\)}
            \State \(M_C \gets f_C(I,\mathcal{T},\mathcal{C})\). \Comment{chart JSX}
        \Else
            \State \(M_C \gets \emptyset\).
        \EndIf
        \State \(P \gets f_W(I,\mathcal{T},\mathcal{C},M_L,M_C)\). \Comment{direct JSX}
        \State \(\hat{I} \gets R(P,W,H)\). \Comment{terminal execution}
        \State \Return \(P,\hat{I}\).
    \end{algorithmic}
\end{algorithm}

\paragraph{Evidence-conditioned layout reasoning.}
The layout reasoner produces a high-level description
\begin{equation}
\label{eq:layout}
M_L = f_L(I,\mathcal{T},\mathcal{C})
\end{equation}
that identifies the main spatial regions and whether the widget contains a chart.
This reasoning operates on the widget as a whole rather than predicting a sequence of components from a predefined vocabulary.

\paragraph{Optional chart reasoning.}
When \(M_L\) indicates a chart, a dedicated call generates chart JSX; otherwise the call is skipped:
\begin{equation}
\label{eq:chart}
M_C =
\begin{cases}
f_C(I,\mathcal{T},\mathcal{C}), & \mathbb{I}_{\mathtt{chart}}(M_L)=1,\\
\emptyset, & \text{otherwise}.
\end{cases}
\end{equation}
When present, the resulting chart code is supplied as context to the final generator.
Separating this call focuses generative capacity on chart geometry only when needed.

\paragraph{Direct JSX synthesis.}
The final generator jointly receives the screenshot, extracted evidence, layout description, and optional chart code:
\begin{equation}
\label{eq:widget}
P = f_W(I,\mathcal{T},\mathcal{C},M_L,M_C).
\end{equation}
It produces executable JSX directly, without translating through component templates or a customized intermediate UI language.

\paragraph{Terminal execution.}
The browser renderer executes \(P\) at the observed dimensions,
\begin{equation}
\label{eq:render}
\hat{I}=R(P,W,H).
\end{equation}
The renderer verifies that the generated artifact can be executed and produces the image used for evaluation.

\paragraph{Learning from executable reconstructions.}
\label{sec:sft_method}
Each successfully executed reconstruction yields an aligned image--code pair for supervised fine-tuning.
Given a generated program \(P_i\), browser execution produces \(\hat{I}_i=R(P_i,W_i,H_i)\); the rendered image, its dimensions, and the OCR and palette evidence recomputed from it form the model input, while \(P_i\) is the target JSX.
Because the target program is executable by construction, these pairs require no manual code annotation and have an unambiguous correspondence between image and code.
Moreover, since the programs originate from reconstructing visual designs rather than unconstrained code synthesis, the resulting pairs retain layout and appearance patterns from the target task.
We use them to fine-tune open-weight MLLMs for evidence-conditioned JSX generation.

% !TEX root = ../main.tex
\section{Experiments}
\label{sec:exp}

\newcommand{\pendingresult}{\textbf{TBD}}

We evaluate WidgetGen along five dimensions: reconstruction fidelity, inference efficiency, the contribution of individual evidence and reasoning stages, qualitative behavior on representative widgets, and the utility of reconstruction-derived supervision for open-weight models.

%%%%%%%%%%%%%%%%%%%%%%%
\subsection{Experimental Setup}
\label{sec:setup}

\paragraph{Benchmark.}
We use the \emph{widget2code-benchmark} introduced by~\cite{widgetfactory}, a collection of widget screenshots for the task introduced in \Cref{sec:task}.
The benchmark is split into a train set of \(1{,}822\) widgets and a test set of \(1{,}000\) widgets.

\paragraph{Evaluation metrics.}
We evaluate every generated widget by rendering its JSX program back to a PNG and comparing the rendered image against the ground-truth screenshot.
We evaluate reconstruction using nine metrics adopted from Widget2Code~\cite{widgetfactory}.
The metrics are computed only on rendered and ground-truth PNGs, which suits this image-only benchmark and applies uniformly to all methods.
They cover five groups: layout, measured by Content Aspect Ratio Similarity (\textbf{Content}) and Area Ratio Similarity (\textbf{Area}); legibility, measured by OCR Text Jaccard (\textbf{Text}) and Local Contrast Similarity (\textbf{LocCon}); style, measured by Palette Distance (\textbf{Palette}) and Vibrancy Consistency (\textbf{Vibrancy}); perceptual similarity, measured by \textbf{SSIM}~\cite{ssim} and \textbf{LPIPS}~\cite{lpips}; and \textbf{Geometry}, which compares the overall aspect ratio and normalized dimensions of the generated and target widgets.
Layout, legibility, style, and geometry metrics are reported on \([0,100]\), where larger is better; SSIM is larger-is-better on \([0,1]\), and LPIPS is smaller-is-better.

\paragraph{Compared methods and backbones.}
We compare three methods for generating UI code from visual widgets, using the same six multimodal models for each method.
\emph{Single-shot Prompting} sends the target widget screenshot with one instruction and asks the model to write JSX in one response.
\emph{Widget2Code}~\cite{widgetfactory} writes a widget description in a custom intermediate language, compiles it into front-end code, and uses adaptive rendering to refine dimensions.
\emph{WidgetGen (ours)} is the tool-grounded framework described in \Cref{sec:overview}.
We use GPT-4o~\cite{gpt4o}, Gemini 3 Pro~\cite{gemini3pro}, Claude Opus 4.5~\cite{claudeopus45}, Seed 2.0 Pro~\cite{seed2}, Qwen3-VL-Plus~\cite{qwen3vlplus}, and Qwen3.5-Plus~\cite{qwen35plus}.
Across all six configurations, only the language model changes; the tools and renderer remain the same.

\paragraph{Implementation details.}
We extract text evidence using EasyOCR on a single H200 GPU\@.
For palette evidence, we apply octree color quantization with 16 initial clusters, merge visually similar colors, discard colors covering less than \(3\%\) of the image, and retain up to eight colors ranked by coverage.
The renderer uses a four-page browser worker pool and executes each generated widget at its target dimensions.

%%%%%%%%%%%%%%%%%%%%%%%
\subsection{Main Results: Reconstruction Fidelity}
\label{sec:main}

\Cref{tab:main_results} reports the nine reconstruction metrics for the three compared methods on the \(1{,}000\)-widget held-out test split.
For each model, the best result on each metric is shown in bold and the second-best result is underlined.

\begin{table}[htbp]
\centering
\caption{Reconstruction fidelity on the \(1{,}000\)-widget held-out test split. Each model block compares the same multimodal backbone under single-shot prompting, Widget2Code~\cite{widgetfactory}, and WidgetGen. The best result within each block is bold and the second best is underlined.}
\label{tab:main_results}
\setlength{\tabcolsep}{3pt}
\resizebox{\textwidth}{!}{%
\begin{tabular}{llccccccccc}
\toprule
& & \multicolumn{2}{c|}{Layout (\(\uparrow\))} & \multicolumn{2}{c|}{Legibility (\(\uparrow\))} & \multicolumn{2}{c|}{Style (\(\uparrow\))} & \multicolumn{2}{c|}{Perceptual} & Geometry \\
\cmidrule(lr){3-4}\cmidrule(lr){5-6}\cmidrule(lr){7-8}\cmidrule(lr){9-10}\cmidrule(lr){11-11}
Model & Method & Content & Area & Text & LocCon & Palette & Vibrancy & SSIM\(\uparrow\) & LPIPS\(\downarrow\) & Geometry\(\uparrow\) \\
\midrule
\multirow{3}{*}{Qwen3-VL-Plus}
 & Single-shot Prompting     & 16.65 & 66.22 & 58.32 & 58.59 & 44.30 & 39.49 & 0.61 & 0.44 & 84.65 \\
 & Widget2Code               & \underline{26.32} & \underline{76.26} & \underline{69.27} & \underline{64.72} & \underline{57.28} & \underline{50.12} & \underline{0.70} & \underline{0.35} & \underline{98.00} \\
 & WidgetGen (ours)           & \textbf{27.60} & \textbf{85.94} & \textbf{72.20} & \textbf{70.55} & \textbf{61.62} & \textbf{58.86} & \textbf{0.710} & \textbf{0.338} & \textbf{100.00} \\
 
\midrule
\multirow{3}{*}{Qwen3.5-Plus}
 & Single-shot Prompting     & 26.10 & 72.25 & 56.74 & 53.39 & 43.78 & 40.98 & 0.58 & 0.42 & 68.78 \\
 & Widget2Code               & \textbf{35.01} & \underline{82.03} & \underline{69.72} & \underline{69.54} & \underline{58.79} & \underline{50.96} & \underline{0.70} & \underline{0.34} & \underline{99.40} \\
 & WidgetGen (ours)          & \underline{31.86} & \textbf{87.61} & \textbf{73.79} & \textbf{73.98} & \textbf{63.86} & \textbf{62.03} & \textbf{0.72} & \textbf{0.32} & \textbf{100.00} \\
\midrule
\multirow{3}{*}{Gemini 3 Pro}
 & Single-shot Prompting     & 38.67 & 83.99 & 69.32 & 77.39 & 57.05 & \underline{56.27} & 0.70 & \underline{0.31} & 95.40 \\
 & Widget2Code               & \underline{40.26} & \underline{85.30} & \underline{70.75} & \underline{77.45} & \underline{62.27} & 55.52 & \underline{0.72} & 0.32 & \underline{99.90} \\
 & WidgetGen (ours)          & \textbf{42.79} & \textbf{88.97} & \textbf{82.46} & \textbf{77.71} & \textbf{68.09} & \textbf{69.97} & \textbf{0.75} & \textbf{0.26} & \textbf{100.00} \\
\midrule
\multirow{3}{*}{Claude Opus 4.5}
 & Single-shot Prompting     & 29.18 & 80.54 & \underline{73.59} & 62.35 & 46.89 & 44.53 & \underline{0.70} & \textbf{0.32} & 98.90 \\
 & Widget2Code               & \textbf{35.79} & \underline{82.96} & 69.22 & \underline{67.49} & \underline{54.41} & \underline{49.60} & 0.69 & \underline{0.35} & \underline{99.50} \\
 & WidgetGen (ours)          & \underline{33.97} & \textbf{88.49} & \textbf{74.80} & \textbf{71.77} & \textbf{62.84} & \textbf{61.68} & \textbf{0.72} & \textbf{0.32} & \textbf{100.00} \\
\midrule
\multirow{3}{*}{GPT-4o}
 & Single-shot Prompting     & 18.08 & 56.26 & 60.89 & 52.55 & 39.96 & 35.82 & \underline{0.63} & \underline{0.40} & 90.55 \\
 & Widget2Code               & \underline{25.16} & \underline{76.17} & \underline{67.72} & \underline{56.94} & \underline{55.65} & \underline{49.59} & \textbf{0.71} & \textbf{0.34} & \underline{99.90} \\
 & WidgetGen (ours)          & \textbf{25.27} & \textbf{85.56} & \textbf{74.32} & \textbf{69.60} & \textbf{60.36} & \textbf{56.16} & \textbf{0.71} & \textbf{0.34} & \textbf{100.00} \\
\midrule
\multirow{3}{*}{Seed 2.0 Pro}
 & Single-shot Prompting     & 26.23 & 73.21 & 56.22 & 61.80 & 48.40 & 44.00 & 0.60 & 0.44 & 85.93 \\
 & Widget2Code               & \underline{34.76} & \underline{82.13} & \underline{64.25} & \underline{72.30} & \underline{58.41} & \underline{50.28} & \underline{0.67} & \underline{0.37} & \underline{99.80} \\
 & WidgetGen (ours)          & \textbf{35.36} & \textbf{88.24} & \textbf{72.75} & \textbf{73.64} & \textbf{64.65} & \textbf{62.81} & \textbf{0.71} & \textbf{0.32} & \textbf{100.00} \\
\bottomrule
\end{tabular}
}
\end{table}

\Cref{tab:main_results} shows three main patterns.
WidgetGen improves legibility and style over both baselines for every model, reaches a geometry score of \(100.00\) for every model, and obtains the best Area score for all six models.
It also obtains the best Content score for four of six models, with Widget2Code higher only for Qwen3.5-Plus and Claude Opus 4.5.
Geometry is nearly saturated for both structured methods and is therefore not the main source of separation.
The metric pattern is consistent with their different control mechanisms: Widget2Code constrains generation through a DSL and compiler, whereas WidgetGen supplies measured text and color evidence before direct JSX synthesis.
The Content exceptions indicate that representation-level structure remains useful for some backbones; the ablations in \Cref{sec:ablation} are designed to separate this effect from additional reasoning and evidence extraction.

%%%%%%%%%%%%%%%%%%%%%%%
\subsection{Fidelity--Efficiency Trade-off}
\label{sec:efficiency}

Architectural simplicity does not by itself establish computational efficiency.
We therefore profile all methods with the same backbone and benchmark, recording inference usage, specialized-tool overhead, end-to-end latency, and rendering reliability.
To distinguish evidence grounding from simply using more model calls, the comparison also includes a call-matched multi-stage baseline that performs layout and chart reasoning without OCR or palette evidence.

\begin{table}[t]
\centering
\caption{Inference-budget, execution-reliability, and visual-fidelity measurements for Qwen3-VL-Plus.}
\label{tab:efficiency}
\setlength{\tabcolsep}{5pt}
\resizebox{\textwidth}{!}{%
\begin{tabular}{lcccc}
\toprule
Measurement & Single-shot & Multi-stage, no tools & Widget2Code & WidgetGen \\
\midrule
MLLM calls / widget & 1.00 & 2.27 & 3.47 & 2.27 \\
Input tokens / widget & 807 & 14{,}593 & 39{,}982 & 14{,}909 \\
Output tokens / widget & 1{,}202 & 2{,}595 & 2{,}035 & 2{,}607 \\
Total tokens / widget (input + output) & 2{,}009 & 17{,}188 & 42{,}017 & 17{,}516 \\
Specialized-tool time (s) & 0 & 0 & 24.5 & 3.88 \\
End-to-end latency, p50 / p95 (s) & 31.5 / 81.3 & 41.8 / 98.8 & 112 / 363 & 98.9 / 271.3 \\
API cost / 1{,}000 widgets (USD) & 1.84 & 5.81 & 8.38 & 5.87 \\
Render success (\%) & 86.0 & 98.3 & 97.0 & 98.0 \\
SSIM $\uparrow$ & 0.61 & 0.69 & 0.70 & 0.71 \\
LPIPS $\downarrow$ & 0.44 & 0.35 & 0.35 & 0.34 \\
\bottomrule
\end{tabular}
}
\end{table}

Palette extraction and OCR run in parallel; the reported WidgetGen tool time is the maximum wall-clock time of the two tools, while the single-shot and no-tools variants do not run either tool.
API cost is computed using the official model pricing and reported in U.S. dollars.
We report fidelity as a function of total MLLM tokens, monetary cost, and latency.
Overall, WidgetGen provides a favorable fidelity--efficiency trade-off, improving reconstruction quality while reducing inference cost relative to the structured pipeline.

%%%%%%%%%%%%%%%%%%%%%%%
\subsection{Evidence Ablation}
\label{sec:ablation}

We isolate the effects of extracted evidence and high-level reasoning using the same model and evaluation setup.
The call-matched no-tools variant retains layout and optional chart reasoning but removes OCR and palette evidence; the remaining variants add one evidence source at a time or remove chart reasoning from the full system.
Chart reasoning is additionally evaluated on the chart-containing subset so that its effect is not diluted by widgets for which the call is skipped.

\begin{table}[t]
\centering
\caption{Ablation of evidence and reasoning stages with Qwen3-VL-Plus. O, P, L, and C denote OCR evidence, palette evidence, layout reasoning, and chart reasoning, respectively. The \emph{All} rows use the full test split; the \emph{Charts} rows use only chart-containing widgets. Within each subset, the best result is bold and the second best is underlined.}
\label{tab:ablation}
\setlength{\tabcolsep}{2.5pt}
\resizebox{\textwidth}{!}{%

\begin{tabular}{ll|cccc|cc|cc|cc|cc|c}
\toprule
& & \multicolumn{4}{c|}{Modules}
& \multicolumn{2}{c|}{Layout (\(\uparrow\))}
& \multicolumn{2}{c|}{Legibility (\(\uparrow\))}
& \multicolumn{2}{c|}{Style (\(\uparrow\))}
& \multicolumn{2}{c|}{Perceptual}
& Geometry \\
\cmidrule(lr){3-6}
\cmidrule(lr){7-8}
\cmidrule(lr){9-10}
\cmidrule(lr){11-12}
\cmidrule(lr){13-14}
\cmidrule(lr){15-15}
Subset & Variant & O & P & L & C
& Content & Area
& Text & LocCon
& Palette & Vibrancy
& SSIM\(\uparrow\) & LPIPS\(\downarrow\)
& Geometry\(\uparrow\) \\
\midrule
\multirow{5}{*}{All}
 & Single-shot
 & -- & -- & -- & --
 & 16.65 & 66.22
 & 58.32 & 58.59
 & 44.30 & 39.49
 & 0.610 & 0.440
 & 84.65 \\

 & Multi-stage, no tools
 & -- & -- & Y & Y
 & 26.58 & 85.42
 & 70.05 & 69.44
 & 61.43 & 58.40
 & 0.690 & 0.350
 & \underline{99.99} \\

 & OCR only
 & Y & -- & Y & Y
 & 25.71 & 85.71
 & \underline{71.58} & 69.62
 & 61.31 & 58.42
 & 0.707 & 0.340
 & \underline{99.99} \\

 & Palette only
 & -- & Y & Y & Y
 & \underline{27.18} & \underline{85.78}
 & 70.00 & \underline{70.18}
 & \underline{61.44} & \textbf{59.05}
 & \underline{0.709} & \underline{0.339}
 & \underline{99.99} \\

 & Full WidgetGen
 & Y & Y & Y & Y
 & \textbf{27.60} & \textbf{85.94}
 & \textbf{72.20} & \textbf{70.55}
 & \textbf{61.62} & \underline{58.86}
 & \textbf{0.710} & \textbf{0.338}
 & \textbf{100.00} \\

\midrule
\multirow{2}{*}{Charts}
 & w/o chart reasoning
 & Y & Y & Y & --
 & \underline{25.59} & \textbf{87.20}
 & \underline{71.09} & \textbf{72.74}
 & \underline{60.21} & \underline{54.92}
 & \underline{0.723} & \underline{0.339}
 & \underline{99.95} \\

 & Full WidgetGen
 & Y & Y & Y & Y
 & \textbf{27.49} & \underline{86.52}
 & \textbf{72.53} & \underline{72.51}
 & \textbf{60.24} & \textbf{56.66}
 & \textbf{0.725} & \textbf{0.336}
 & \textbf{100.00} \\
\bottomrule
\end{tabular}

}
\end{table}

Compared with single-shot prompting, the multi-stage variant without tools improves the reconstruction metrics, indicating that high-level reasoning contributes to reconstruction quality.
Adding text and color evidence produces further gains in legibility and style, with the full configuration achieving the best Text, Palette, and Geometry scores.
On the chart subset, chart reasoning improves several content and appearance metrics.

%%%%%%%%%%%%%%%%%%%%%%%
\subsection{Qualitative Analysis}
\label{sec:qualitative}

\begin{figure}[t]
\centering
\setlength{\tabcolsep}{1pt}
\begin{tabular}{cccc}
Target & Single-shot & Widget2Code & WidgetGen \\
\includegraphics[width=0.245\textwidth]{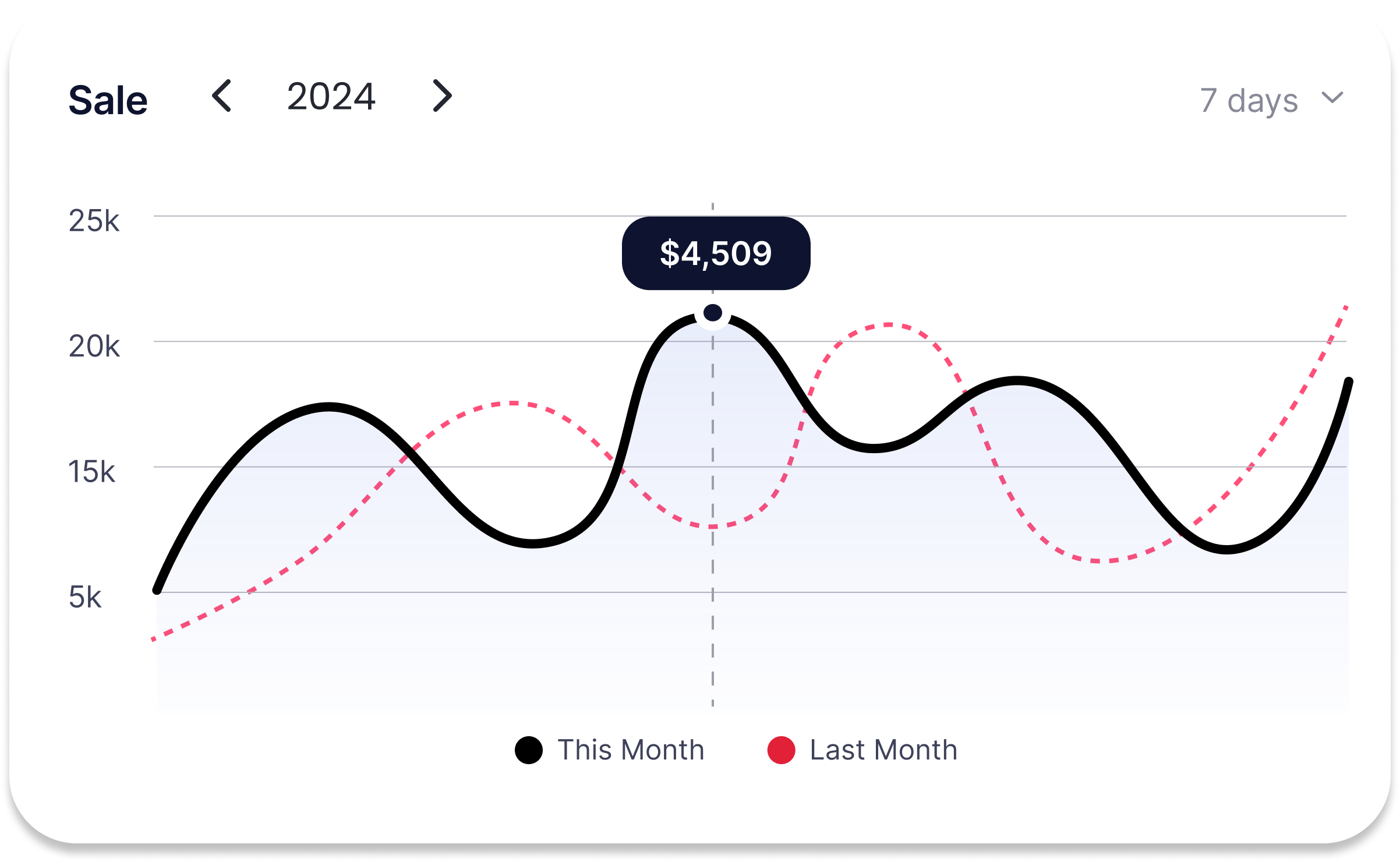} &
\includegraphics[width=0.245\textwidth]{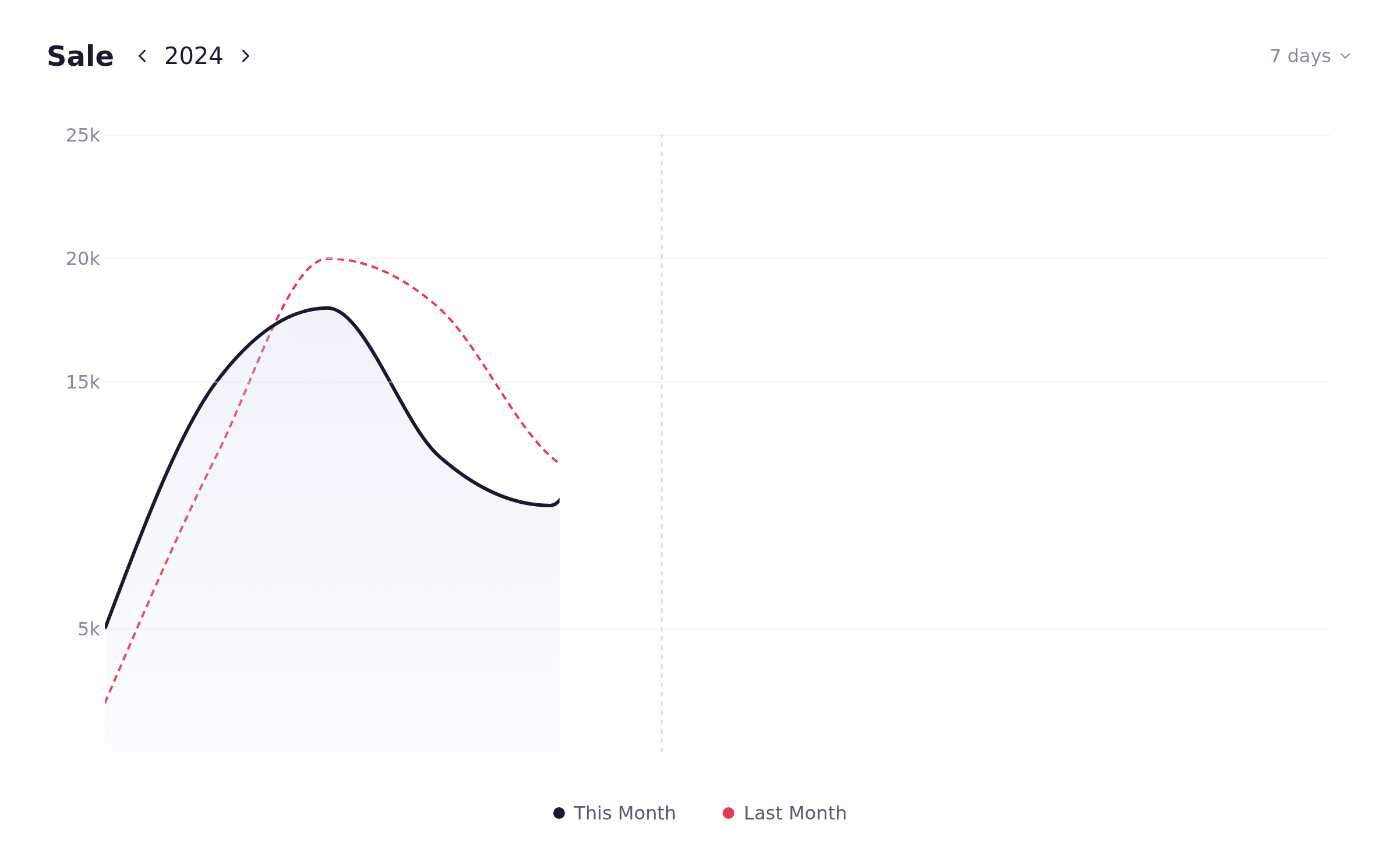} &
\includegraphics[width=0.245\textwidth]{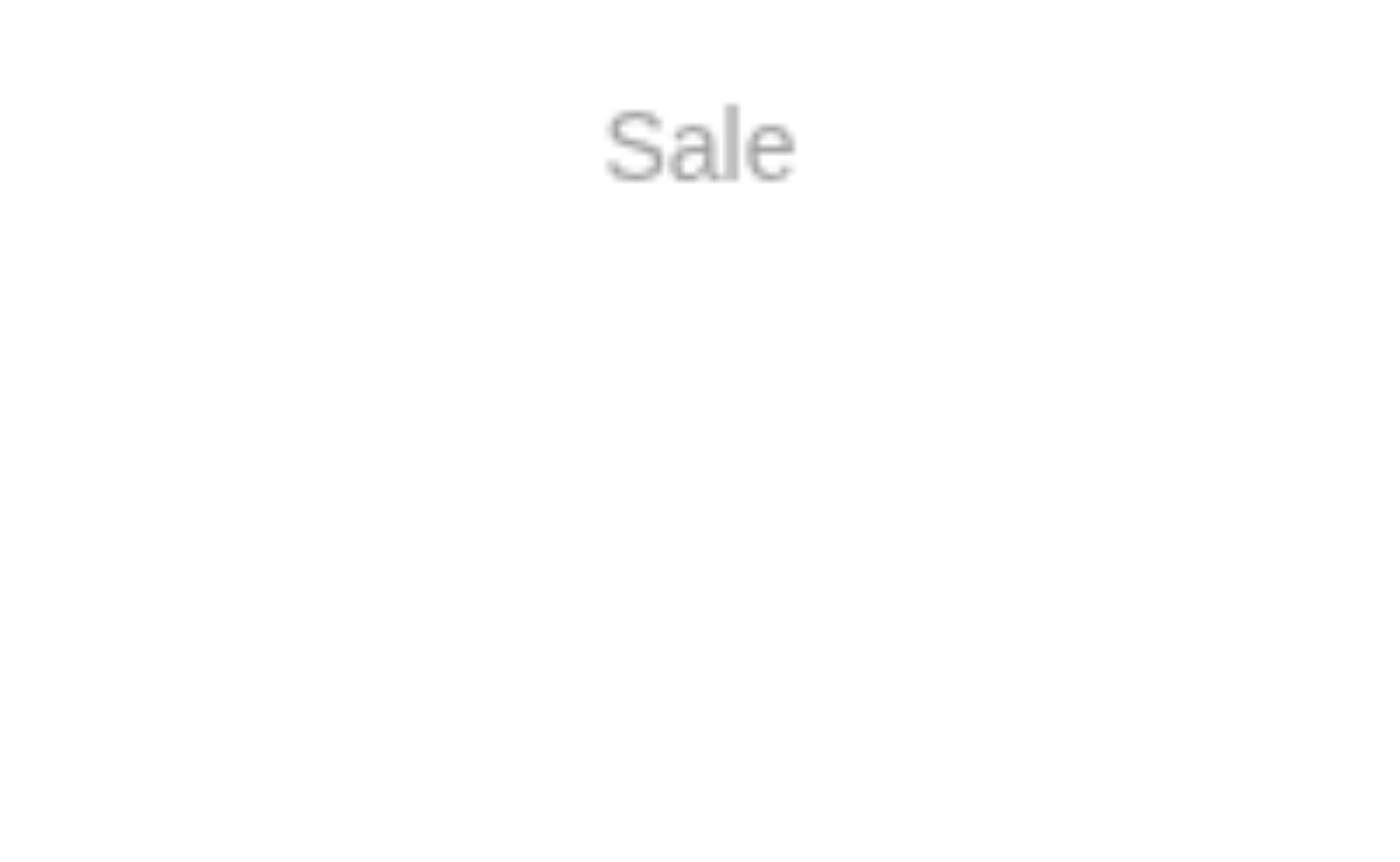} &
\includegraphics[width=0.245\textwidth]{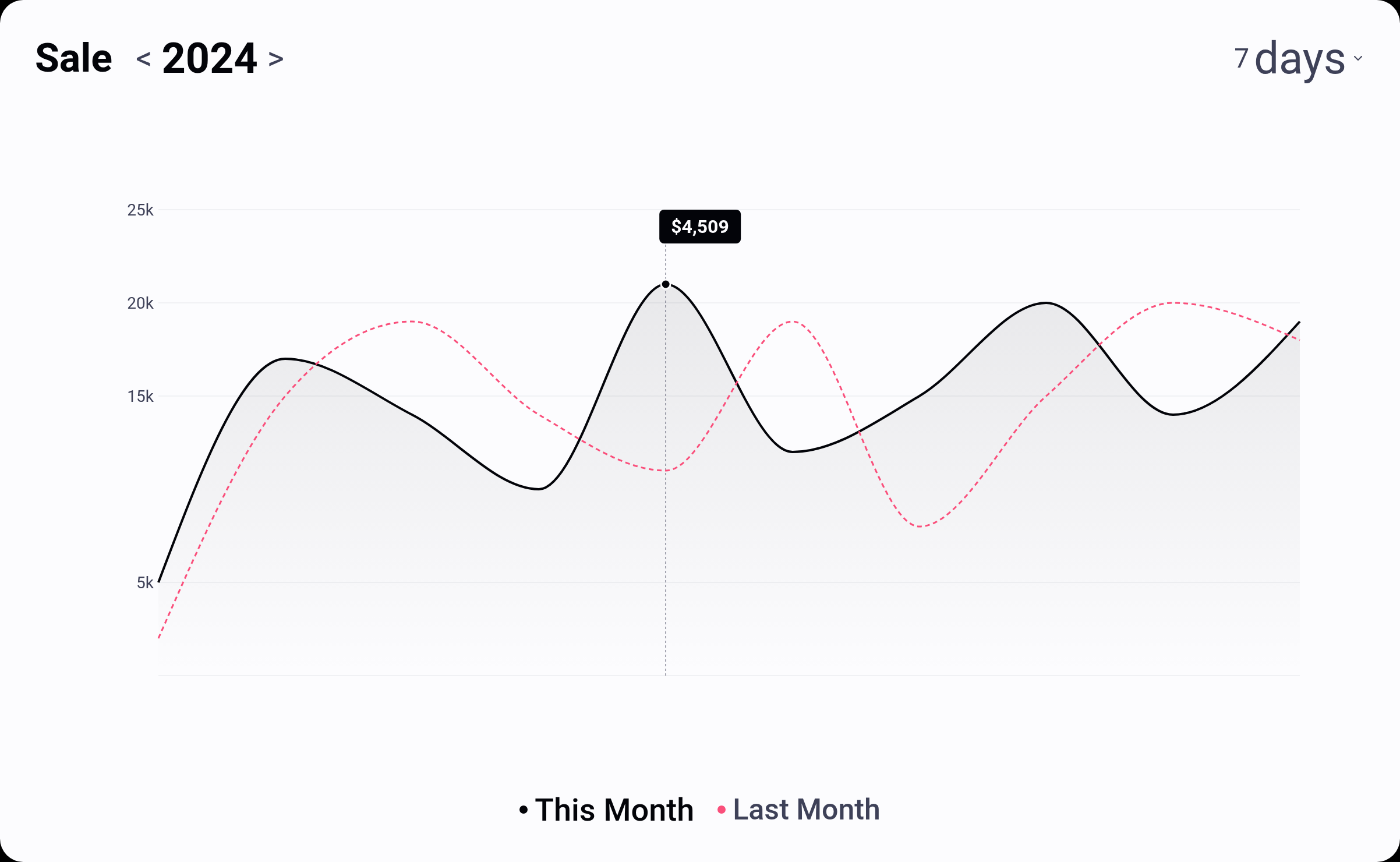} \\
\includegraphics[width=0.245\textwidth]{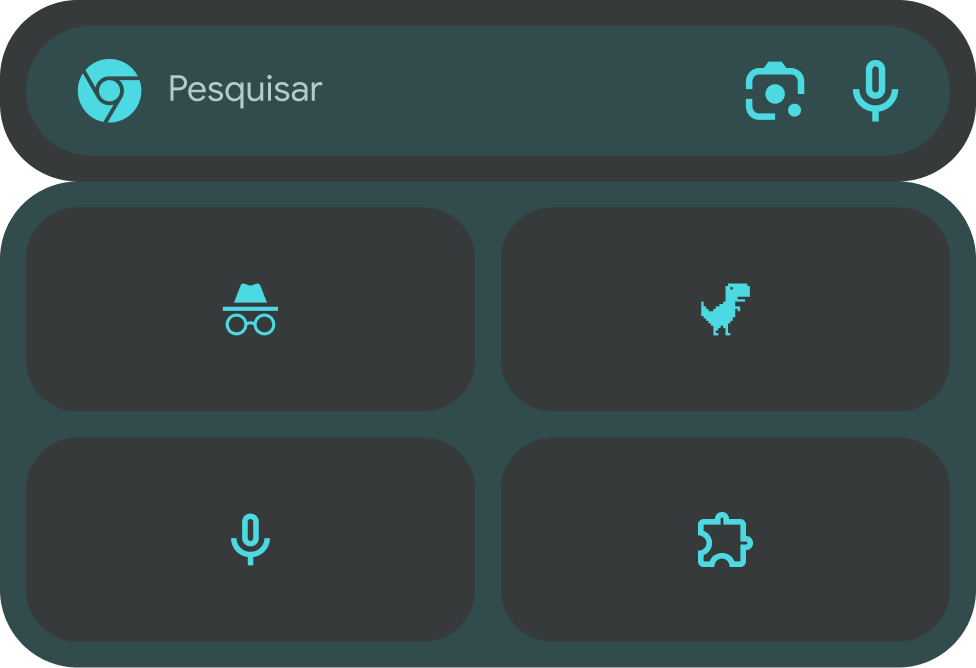} &
\includegraphics[width=0.245\textwidth]{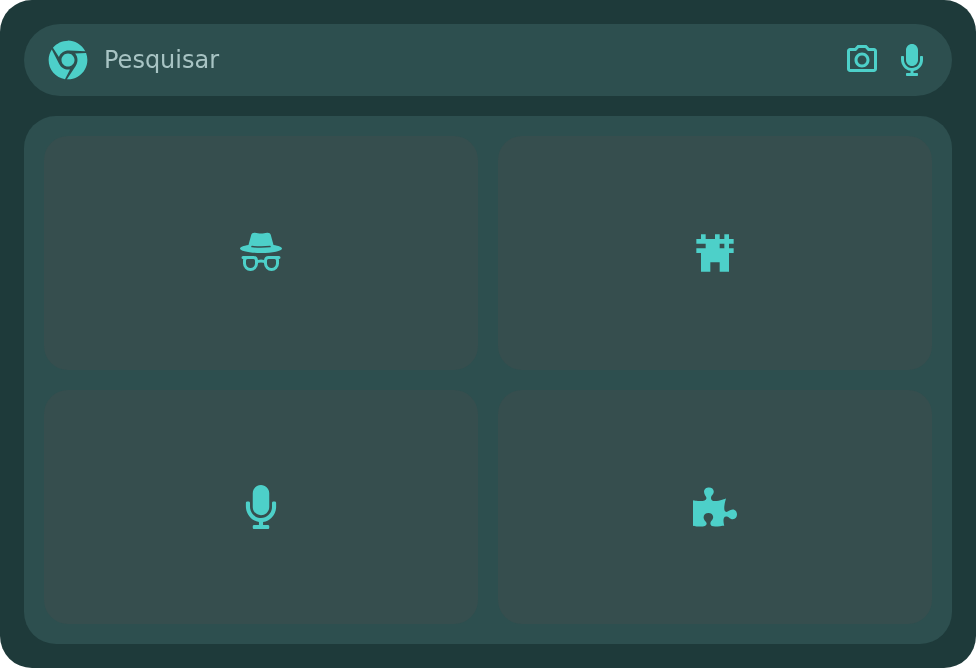} &
\includegraphics[width=0.245\textwidth]{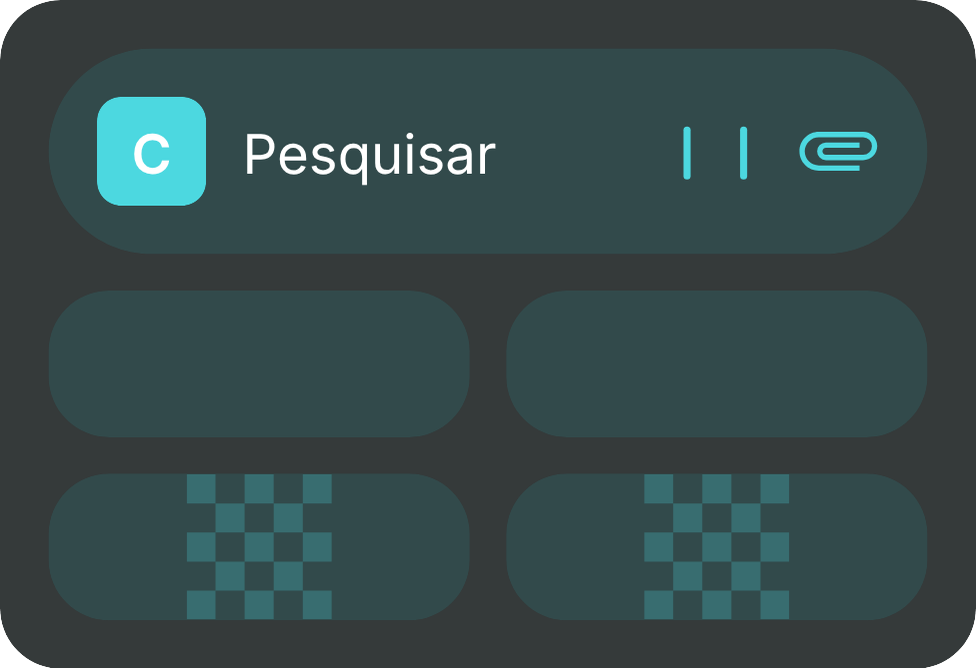} &
\includegraphics[width=0.245\textwidth]{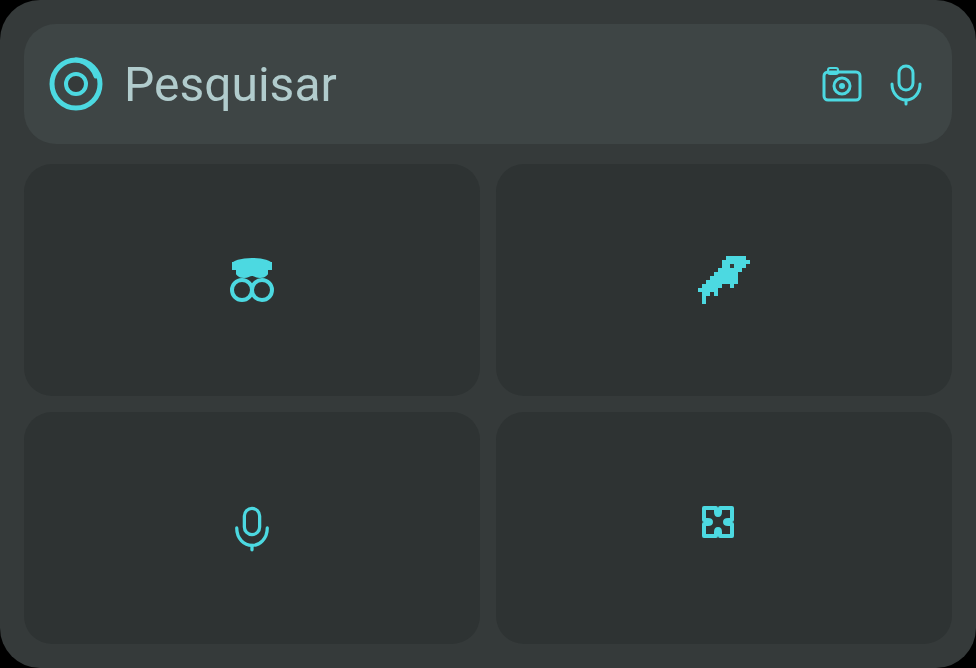} \\
\includegraphics[width=0.245\textwidth]{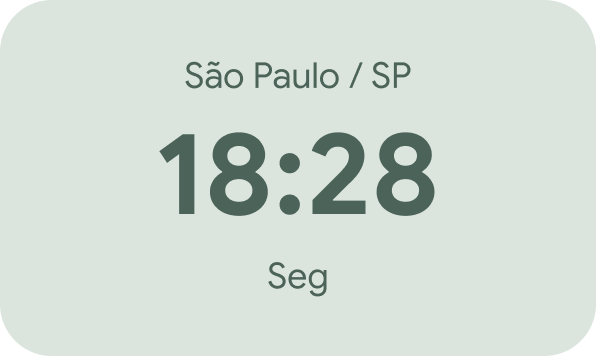} &
\includegraphics[width=0.245\textwidth]{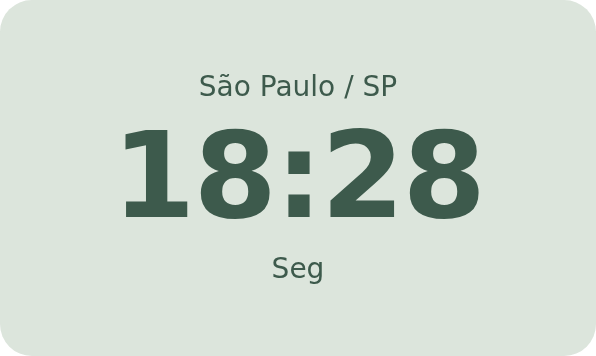} &
\includegraphics[width=0.245\textwidth]{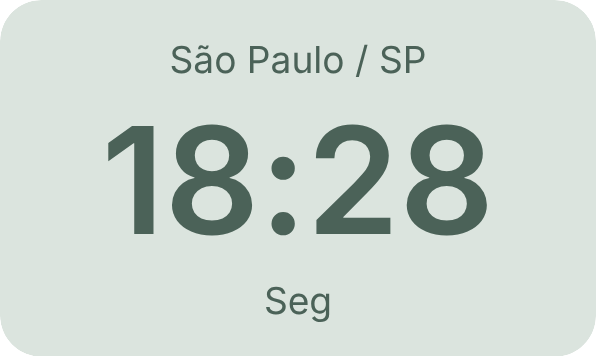} &
\includegraphics[width=0.245\textwidth]{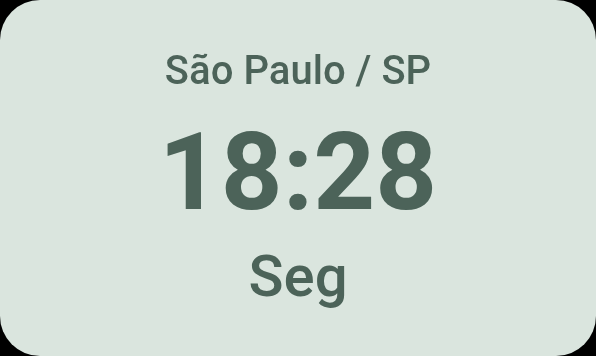}
\end{tabular}
\caption{Qualitative comparison on chart reconstruction (top), icon-dense composition (middle), and compact typography (bottom). The examples include both clear improvements and residual errors in chart shape, icon identity, and text styling.}
\label{fig:qualitative}
\end{figure}

\paragraph{Representative reconstructions.}
\Cref{fig:qualitative} compares the three methods on a chart, an icon-dense grid, and a compact text widget.
In the chart example, single-shot prompting reconstructs only part of the plot and Widget2Code omits most chart content, whereas WidgetGen recovers the principal curves, axes, legend, and annotation while retaining visible shape differences from the target.
The icon-grid example shows improved preservation of the global composition and palette, but also exposes residual errors in icon identity.
In the compact time widget, WidgetGen more closely matches the target's typographic weight and text placement while preserving the overall composition.

%%%%%%%%%%%%%%%%%%%%%%%
\subsection{Generated Supervision for Open-Weight Models}
\label{sec:sft}

As a secondary study, we evaluate whether the executable reconstructions described in \Cref{sec:sft_method} provide useful task-specific supervision for open-weight vision-language models.
This experiment is separate from the main system comparison: it studies WidgetGen as a data-generation process rather than as an inference-time replacement for the base model.

\begin{table}[htbp]
\centering
\caption[Generated-supervision results.]{Effect of reconstruction-derived supervision on the held-out test set. Rows without \texttt{-SFT} are the original checkpoints; rows ending in \texttt{-SFT} are adapted using WidgetGen-generated pairs.}
\label{tab:sft_results}
\setlength{\tabcolsep}{3pt}
\resizebox{\textwidth}{!}{%
\begin{tabular}{lccccccccc}
\toprule
& \multicolumn{2}{c|}{Layout (\(\uparrow\))} & \multicolumn{2}{c|}{Legibility (\(\uparrow\))} & \multicolumn{2}{c|}{Style (\(\uparrow\))} & \multicolumn{2}{c|}{Perceptual} & Geometry \\
\cmidrule(lr){2-3}\cmidrule(lr){4-5}\cmidrule(lr){6-7}\cmidrule(lr){8-9}\cmidrule(lr){10-10}
Model & Content & Area & Text & LocCon & Palette & Vibrancy & SSIM\(\uparrow\) & LPIPS\(\downarrow\) & Geometry\(\uparrow\) \\
\midrule
Qwen3-VL-4B-Instruct & 19.09 & 53.24 & 51.26 & 41.49 & 43.51 & 39.57 & 0.64 & 0.36 & 60.93 \\
Qwen3-VL-4B-Instruct-SFT & \textbf{26.01} & \textbf{80.03} & \textbf{66.01} & \textbf{70.10} & \textbf{60.28} & \textbf{56.70} & \textbf{0.70} & \textbf{0.34} & \textbf{100.00} \\
\midrule
Qwen3-VL-8B-Instruct & 22.72 & 58.05 & 58.12 & 43.84 & 44.90 & 42.47 & 0.70 & 0.34 & 65.79 \\
Qwen3-VL-8B-Instruct-SFT & \textbf{29.97} & \textbf{80.99} & \textbf{68.80} & \textbf{70.43} & \textbf{59.73} & \textbf{57.83} & \textbf{0.71} & \textbf{0.33} & \textbf{100.00} \\
\midrule
Qwen3-VL-32B-Instruct & 17.03 & 57.34 & 66.14 & 58.34 & 47.21 & 43.10 & 0.70 & 0.34 & 95.24 \\
Qwen3-VL-32B-Instruct-SFT & \textbf{34.85} & \textbf{84.86} & \textbf{73.51} & \textbf{73.17} & \textbf{62.95} & \textbf{61.11} & \textbf{0.72} & \textbf{0.31} & \textbf{99.71} \\
\midrule
Qwen3.5-9B & 20.15 & 64.34 & 55.60 & 44.57 & 46.58 & 42.80 & 0.69 & 0.35 & 55.57 \\
Qwen3.5-9B-SFT & \textbf{36.85} & \textbf{86.33} & \textbf{72.89} & \textbf{73.19} & \textbf{64.37} & \textbf{63.82} & \textbf{0.73} & \textbf{0.29} & \textbf{100.00} \\
\midrule
Qwen3.5-27B & 27.68 & 80.65 & 63.87 & 51.13 & 50.39 & 47.32 & 0.67 & 0.35 & 55.95 \\
Qwen3.5-27B-SFT & \textbf{41.92} & \textbf{87.91} & \textbf{75.89} & \textbf{76.75} & \textbf{65.90} & \textbf{66.36} & \textbf{0.74} & \textbf{0.27} & \textbf{100.00} \\
\midrule
Qwen3.6-27B & 30.53 & 82.22 & 65.41 & 49.54 & 52.01 & 47.47 & 0.69 & 0.34 & 50.35 \\
Qwen3.6-27B-SFT & \textbf{42.80} & \textbf{88.84} & \textbf{75.67} & \textbf{76.20} & \textbf{67.30} & \textbf{68.04} & \textbf{0.74} & \textbf{0.26} & \textbf{100.00} \\
\bottomrule
\end{tabular}
}
\end{table}

\begin{table}[t]
\centering
\caption{Supervision-route comparison on Qwen3-VL-32B-Instruct using \(1{,}822\) training pairs per route and the same nine metrics as the main evaluation. The best result is bold and the second best is underlined.}
\label{tab:data_ablation}
\setlength{\tabcolsep}{5pt}
\resizebox{\textwidth}{!}{%
\begin{tabular}{l|cc|cc|cc|cc|c}
\toprule
& \multicolumn{2}{c|}{Layout (\(\uparrow\))} & \multicolumn{2}{c|}{Legibility (\(\uparrow\))} & \multicolumn{2}{c|}{Style (\(\uparrow\))} & \multicolumn{2}{c|}{Perceptual} & Geometry \\
\cmidrule(lr){2-3}\cmidrule(lr){4-5}\cmidrule(lr){6-7}\cmidrule(lr){8-9}\cmidrule(lr){10-10}
Supervision & Content & Area & Text & LocCon & Palette & Vibrancy & SSIM\(\uparrow\) & LPIPS\(\downarrow\) & Geometry\(\uparrow\) \\
\midrule
None (base checkpoint) & 17.03 & 57.34 & 66.14 & 58.34 & 47.21 & 43.10 & 0.70 & \underline{0.34} & 95.24 \\
WidgetFactory-generated pairs & \underline{24.15} & \underline{75.81} & \underline{68.95} & \underline{62.43} & \underline{52.61} & \underline{46.13} & \underline{0.709} & 0.343 & \textbf{100.00} \\
WidgetGen reconstruction pairs & \textbf{34.85} & \textbf{84.86} & \textbf{73.51} & \textbf{73.17} & \textbf{62.95} & \textbf{61.11} & \textbf{0.72} & \textbf{0.31} & \underline{99.71} \\
\bottomrule
\end{tabular}
}
\end{table}

\paragraph{Training data.}
We apply WidgetGen with Gemini 3.1 Pro~\cite{gemini31pro} to the \(1{,}822\) training widgets.
Each executable program is rendered to produce its paired image, from which text and color evidence are re-extracted.
Pairing a program with its own rendering preserves image--code correspondence even when the reconstruction differs from the source screenshot.
After quality filtering and regeneration, we obtain \(1{,}822\) executable image--code pairs, one for each training widget.

\paragraph{Training setup.}
We evaluate six checkpoints from the Qwen3-VL~\cite{qwen3vl}, Qwen3.5~\cite{qwen35}, and Qwen3.6~\cite{qwen36} families, ranging from \(4\)B to \(32\)B parameters.
Each model is adapted with LoRA~\cite{lora} while keeping the vision encoder frozen.
We train rank-\(32\) adapters for four epochs with a scaling factor of \(64\), dropout \(0.05\), learning rate \(10^{-4}\), and effective batch size \(16\).
Base and adapted checkpoints use the same evidence-conditioned inputs and decoding setup.

\Cref{tab:sft_results} compares the six checkpoints before and after adaptation using the same nine reconstruction metrics.
For each model, the base row is the original checkpoint and the \texttt{-SFT} row is its adapted counterpart.

\Cref{tab:sft_results} shows consistent improvements across all six checkpoints.
Reconstruction-derived supervision improves every reported metric, with gains spanning layout, legibility, style, and perceptual similarity.
Within this evaluation, the adapted \(27\)B checkpoints also exceed the proprietary single-shot baselines on most metrics. The process is similar to distilling the knowledge~\cite{hinton2015distilling,zhong2022meta,hamidi2024train,ye2024bayes} from stronger Gemini to weaker models.

\Cref{tab:data_ablation} further compares three supervision routes for Qwen3-VL-32B-Instruct: no task-specific training, size-matched WidgetFactory-generated data, and WidgetGen reconstruction pairs.

WidgetGen supervision achieves the best result on eight of the nine metrics.
The comparison indicates that reconstruction-derived JSX pairs provide more effective supervision for the evaluated direct-generation setting than the size-matched DSL-based data.

\section{Conclusion}

We introduced WidgetGen, a lightweight tool-grounded framework for visual widget-to-code generation.
WidgetGen combines extracted text and color evidence with high-level layout and optional chart reasoning before directly synthesizing executable JSX.
Across six multimodal models and \(1{,}000\) held-out widgets, WidgetGen improves most reconstruction metrics over both direct prompting and Widget2Code while providing a favorable fidelity--efficiency trade-off.
Ablations show complementary contributions from high-level reasoning and visual evidence.
In addition, reconstruction-derived image--code pairs improve six open-weight Qwen-family models across all reported metrics.
These results establish WidgetGen as a strong and efficient baseline for widget-to-code generation and demonstrate the value of grounding direct code synthesis with task-relevant visual evidence.

\paragraph{Limitations.}
The current evaluation focuses on static widgets from a single dataset, so generalization to broader UI distributions remains to be established.
Our metrics assess rendered visual fidelity rather than interactive behavior, accessibility, or code maintainability.
The generated-supervision study is limited to Qwen-family models; extending it to other model families remains future work.

\bibliographystyle{splncs04}
\bibliography{references}

@String(CVPR  = {IEEE Conf. Comput. Vis. Pattern Recog.})

@String(ICML  = {Int. Conf. Mach. Learn.})

@String(ICLR  = {Int. Conf. Learn. Represent.})

@String(AAAI  = {AAAI})

@String(CVPR  = {CVPR})

@String(ICML  = {ICML})

@String(ICLR  = {ICLR})

@misc{claudeopus45,
  title={{Claude Opus 4.5}},
  author={{Anthropic}},
  howpublished={System card},
  year={2025},
  note={\url{https://assets.anthropic.com/m/64823ba7485345a7/Claude-Opus-4-5-System-Card.pdf}}
}

@misc{gemini3pro,
  title={{Gemini 3 Pro}},
  author={{Google DeepMind}},
  howpublished={Model card},
  year={2025},
  note={\url{https://storage.googleapis.com/deepmind-media/Model-Cards/Gemini-3-Pro-Model-Card.pdf}}
}

@misc{gemini31pro,
  title={{Gemini 3.1 Pro}},
  author={{Google DeepMind}},
  howpublished={Model card},
  year={2026},
  note={\url{https://deepmind.google/models/model-cards/gemini-3-1-pro/}}
}

@misc{gpt4o,
  title={{GPT-4o} System Card},
  author={{OpenAI}},
  howpublished={System card},
  year={2024},
  note={\url{https://openai.com/index/gpt-4o-system-card/}}
}

@misc{seed2,
  title={{Seed 2.0}},
  author={{ByteDance Seed}},
  howpublished={Model release page},
  year={2026},
  note={\url{https://seed.bytedance.com/en/seed2}}
}

@inproceedings{design2code,
  title={{Design2Code}: Benchmarking Multimodal Code Generation for Automated Front-End Engineering},
  author={Si, Chenglei and Zhang, Yanzhe and Li, Ryan and Yang, Zhengyuan and Liu, Ruibo and Yang, Diyi},
  booktitle={Proceedings of the 2025 Conference of the Nations of the Americas Chapter of the Association for Computational Linguistics: Human Language Technologies (Volume 1: Long Papers)},
  pages={3956--3974},
  year={2025},
  publisher={Association for Computational Linguistics},
  doi={10.18653/v1/2025.naacl-long.199},
  url={https://aclanthology.org/2025.naacl-long.199/}
}

@article{websight,
  title={Unlocking the conversion of Web Screenshots into {HTML} Code with the {WebSight} Dataset},
  author={Lauren{\c{c}}on, Hugo and Tronchon, L{\'e}o and Sanh, Victor},
  journal={arXiv preprint arXiv:2403.09029},
  year={2024}
}

@inproceedings{pix2code,
  title={pix2code: Generating Code from a Graphical User Interface Screenshot},
  author={Beltramelli, Tony},
  booktitle={Proceedings of the ACM SIGCHI Symposium on Engineering Interactive Computing Systems},
  articleno={3},
  pages={3:1--3:6},
  year={2018},
  publisher={ACM},
  doi={10.1145/3220134.3220135}
}

@inproceedings{web2code,
  title={{Web2Code}: A Large-scale Webpage-to-Code Dataset and Evaluation Framework for Multimodal {LLMs}},
  author={Yun, Sukmin and Lin, Haokun and Thushara, Rusiru and Bhat, Mohammad Qazim and Wang, Yongxin and Jiang, Zutao and Deng, Mingkai and Wang, Jinhong and Tao, Tianhua and Li, Junbo and Li, Haonan and Nakov, Preslav and Baldwin, Timothy and Liu, Zhengzhong and Xing, Eric P. and Liang, Xiaodan and Shen, Zhiqiang},
  booktitle={Advances in Neural Information Processing Systems},
  volume={37},
  pages={112134--112157},
  year={2024},
  doi={10.52202/079017-3560}
}

@inproceedings{webcode2m,
  title={{WebCode2M}: A Real-World Dataset for Code Generation from Webpage Designs},
  author={Gui, Yi and Li, Zhen and Wan, Yao and Shi, Yemin and Zhang, Hongyu and Chen, Bohua and Su, Yi and Chen, Dongping and Wu, Siyuan and Zhou, Xing and Jiang, Wenbin and Jin, Hai and Zhang, Xiangliang},
  booktitle={Proceedings of the ACM on Web Conference 2025},
  pages={1834--1845},
  year={2025},
  publisher={Association for Computing Machinery},
  doi={10.1145/3696410.3714889}
}

@inproceedings{waffle,
  title={{WAFFLE}: Fine-tuning Multi-Modal Model for Automated Front-End Development},
  author={Liang, Shanchao and Jiang, Nan and Qian, Shangshu and Tan, Lin},
  booktitle={Proceedings of the 63rd Annual Meeting of the Association for Computational Linguistics (Volume 1: Long Papers)},
  pages={24786--24802},
  year={2025},
  publisher={Association for Computational Linguistics},
  doi={10.18653/v1/2025.acl-long.1208},
  url={https://aclanthology.org/2025.acl-long.1208/}
}

@article{flame,
  title={Advancing Vision-Language Models in Front-End Development via Data Synthesis},
  author={Ge, Tong and Liu, Yashu and Ye, Jieping and Li, Tianyi and Wang, Chao},
  journal={arXiv preprint arXiv:2503.01619},
  year={2025}
}

@inproceedings{webuibench,
  title={{WebUIBench}: A Comprehensive Benchmark for Evaluating Multimodal Large Language Models in {WebUI}-to-Code},
  author={Lin, Zhiyu and Zhou, Zhengda and Zhao, Zhiyuan and Wan, Tianrui and Ma, Yilun and Gao, Junyu and Li, Xuelong},
  booktitle={Findings of the Association for Computational Linguistics: ACL 2025},
  pages={15780--15797},
  year={2025},
  publisher={Association for Computational Linguistics},
  doi={10.18653/v1/2025.findings-acl.815},
  url={https://aclanthology.org/2025.findings-acl.815/}
}

@inproceedings{chartmimic,
  title={{ChartMimic}: Evaluating {LMM}'s Cross-Modal Reasoning Capability via Chart-to-Code Generation},
  author={Yang, Cheng and Shi, Chufan and Liu, Yaxin and Shui, Bo and Wang, Junjie and Jing, Mohan and Xu, Linran and Zhu, Xinyu and Li, Siheng and Zhang, Yuxiang and Liu, Gongye and Nie, Xiaomei and Cai, Deng and Yang, Yujiu},
  booktitle={The Thirteenth International Conference on Learning Representations},
  year={2025},
  url={https://openreview.net/forum?id=sGpCzsfd1K}
}

@article{dcgen,
  title={Divide-and-Conquer: Generating {UI} Code from Screenshots},
  author={Wan, Yuxuan and Wang, Chaozheng and Dong, Yi and Wang, Wenxuan and Li, Shuqing and Huo, Yintong and Lyu, Michael R.},
  journal={Proceedings of the ACM on Software Engineering},
  volume={2},
  number={FSE},
  pages={2099--2122},
  year={2025},
  doi={10.1145/3729364}
}

@inproceedings{latcoder,
  title={{LaTCoder}: Converting Webpage Design to Code with Layout-as-Thought},
  author={Gui, Yi and Li, Zhen and Zhang, Zhongyi and Wang, Guohao and Lv, Tianpeng and Jiang, Gaoyang and Liu, Yi and Chen, Dongping and Wan, Yao and Zhang, Hongyu and Jiang, Wenbin and Shi, Xuanhua and Jin, Hai},
  booktitle={Proceedings of the 31st ACM SIGKDD Conference on Knowledge Discovery and Data Mining V.2},
  pages={721--732},
  year={2025},
  doi={10.1145/3711896.3737016}
}

@inproceedings{uicopilot,
  title={{UICopilot}: Automating {UI} Synthesis via Hierarchical Code Generation from Webpage Designs},
  author={Gui, Yi and Wan, Yao and Li, Zhen and Zhang, Zhongyi and Chen, Dongping and Zhang, Hongyu and Su, Yi and Chen, Bohua and Zhou, Xing and Jiang, Wenbin and Zhang, Xiangliang},
  booktitle={Proceedings of the ACM on Web Conference 2025},
  pages={1846--1855},
  year={2025},
  publisher={Association for Computing Machinery},
  doi={10.1145/3696410.3714891}
}

@article{ui2code_layout,
  title={{MLLM}-Based {UI2Code} Automation Guided by {UI} Layout Information},
  author={Wu, Fan and Gao, Cuiyun and Li, Shuqing and Wen, Xin-Cheng and Liao, Qing},
  journal={Proceedings of the ACM on Software Engineering},
  volume={2},
  number={ISSTA},
  pages={1123--1145},
  year={2025},
  doi={10.1145/3728925}
}

@inproceedings{uiorchestra,
  title={{UIOrchestra}: Generating High-Fidelity Code from {UI} Designs with a Multi-agent System},
  author={Yue, Chuhuai and Chai, Jiajun and Zhang, Yufei and Ding, Zixiang and Liang, Xihao and Wang, Peixin and Chen, Shihai and Yixuan, Wang and {Wangyanping} and Yin, Guojun and Lin, Wei},
  booktitle={Findings of the Association for Computational Linguistics: EMNLP 2025},
  pages={2769--2782},
  year={2025},
  address={Suzhou, China},
  publisher={Association for Computational Linguistics},
  doi={10.18653/v1/2025.findings-emnlp.150},
  url={https://aclanthology.org/2025.findings-emnlp.150/}
}

@article{designcoder,
  title={{DesignCoder}: Hierarchy-aware and self-correcting {UI} code generation with large language models},
  author={Chen, Yunnong and Yu, Xinyu and Ding, Shixian and Zhang, Yingying and Shi, Chengwei and Du, Jingzhou and Chen, Liuqing},
  journal={Information and Software Technology},
  volume={198},
  pages={108214},
  year={2026},
  doi={10.1016/j.infsof.2026.108214}
}

@inproceedings{ui2coden,
  title={{UI2Code}$^{N}$: {UI}-to-Code Generation as Interactive Visual Optimization},
  author={Yang, Zhen and Hong, Wenyi and Xu, Mingde and Fan, Xinyue and Wang, Weihan and Cheng, Jiele and Gu, Xiaotao and Tang, Jie},
  booktitle={Proceedings of the 43rd International Conference on Machine Learning ({ICML})},
  year={2026},
  url={https://zheny2751-dotcom.github.io/ui2code-n.github.io/}
}

@article{designbench,
  title={{DesignBench}: A Comprehensive Benchmark for {MLLM}-based Front-end Code Generation},
  author={Xiao, Jingyu and Wang, Ming and Lam, Man Ho and Wan, Yuxuan and Liu, Junliang and Huo, Yintong and Lyu, Michael R.},
  journal={arXiv preprint arXiv:2506.06251},
  year={2025}
}

@article{uiug,
  title={{UI-UG}: A Unified {MLLM} for {UI} Understanding and Generation},
  author={Yang, Hao and Qiu, Weijie and Zhang, Ru and Fang, Zhou and Mao, Ruichao and Lin, Xiaoyu and Huang, Maji and Huang, Zhaosong and Guo, Teng and Liu, Shuoyang and Rao, Hai},
  journal={arXiv preprint arXiv:2509.24361},
  year={2025}
}

@inproceedings{widgetfactory,
  title={{Widget2Code}: From Visual Widgets to {UI} Code via Multimodal {LLMs}},
  author={Zhang, Houston H. and Zhang, Tao and Lin, Baoze and Xue, Yuanqi and Zhu, Yincheng and Liu, Huan and Gu, Li and Ye, Linfeng and Wang, Ziqiang and Zuo, Xinxin and Wang, Yang and Yu, Yuanhao and Chi, Zhixiang},
  booktitle={Proceedings of the IEEE/CVF Conference on Computer Vision and Pattern Recognition ({CVPR})},
  pages={20293--20302},
  year={2026}
}

@article{qwen3vl,
  title={{Qwen3-VL} Technical Report},
  author={Bai, Shuai and Cai, Yuxuan and Chen, Ruizhe and Chen, Keqin and Chen, Xionghui and Cheng, Zesen and Deng, Lianghao and Ding, Wei and Gao, Chang and Ge, Chunjiang and Ge, Wenbin and Guo, Zhifang and Huang, Qidong and Huang, Jie and Huang, Fei and Hui, Binyuan and Jiang, Shutong and Li, Zhaohai and Li, Mingsheng and Li, Mei and Li, Kaixin and Lin, Zicheng and Lin, Junyang and others},
  journal={arXiv preprint arXiv:2511.21631},
  year={2025}
}

@misc{qwen3vlplus,
  title={Visual Understanding},
  author={{Alibaba Cloud}},
  howpublished={Model documentation},
  year={2026},
  note={\url{https://www.alibabacloud.com/help/en/model-studio/vision-model}}
}

@misc{qwen35plus,
  title={Alibaba Open-Sources {Qwen3.5}, a Natively Multimodal Model Built for High-Efficiency Inference},
  author={{Alibaba Group}},
  howpublished={Model release},
  year={2026},
  note={\url{https://home.alibabagroup.com/en-US/document-1960233590314762240}}
}

@misc{qwen35,
  title={{Qwen3.5}},
  author={{Qwen Team}},
  howpublished={Model collection},
  year={2026},
  note={\url{https://huggingface.co/collections/Qwen/qwen35}}
}

@misc{qwen36,
  title={{Qwen3.6-27B}: Flagship-Level Coding in a 27B Dense Model},
  author={{Qwen Team}},
  howpublished={Model release},
  month={April},
  year={2026},
  note={\url{https://qwen.ai/blog?id=qwen3.6-27b}}
}

@inproceedings{llava,
  title={Visual Instruction Tuning},
  author={Liu, Haotian and Li, Chunyuan and Wu, Qingyang and Lee, Yong Jae},
  booktitle={Advances in Neural Information Processing Systems},
  volume={36},
  pages={34892--34916},
  year={2023}
}

@inproceedings{blip2,
  title={{BLIP-2}: Bootstrapping Language-Image Pre-training with Frozen Image Encoders and Large Language Models},
  author={Li, Junnan and Li, Dongxu and Savarese, Silvio and Hoi, Steven},
  booktitle={Proceedings of the 40th International Conference on Machine Learning ({ICML})},
  volume={202},
  pages={19730--19742},
  publisher={PMLR},
  year={2023}
}

@inproceedings{pal,
  title={{PAL}: Program-Aided Language Models},
  author={Gao, Luyu and Madaan, Aman and Zhou, Shuyan and Alon, Uri and Liu, Pengfei and Yang, Yiming and Callan, Jamie and Neubig, Graham},
  booktitle={Proceedings of the 40th International Conference on Machine Learning ({ICML})},
  volume={202},
  pages={10764--10799},
  publisher={PMLR},
  year={2023},
  url={https://proceedings.mlr.press/v202/gao23f}
}

@inproceedings{lora,
  title={{LoRA}: Low-Rank Adaptation of Large Language Models},
  author={Hu, Edward J. and Shen, Yelong and Wallis, Phillip and Allen-Zhu, Zeyuan and Li, Yuanzhi and Wang, Shean and Wang, Lu and Chen, Weizhu},
  booktitle={International Conference on Learning Representations ({ICLR})},
  year={2022},
  url={https://openreview.net/forum?id=nZeVKeeFYf9}
}

@article{ssim,
  title={Image Quality Assessment: From Error Visibility to Structural Similarity},
  author={Wang, Zhou and Bovik, Alan C. and Sheikh, Hamid R. and Simoncelli, Eero P.},
  journal={IEEE Transactions on Image Processing},
  volume={13},
  number={4},
  pages={600--612},
  year={2004},
  doi={10.1109/TIP.2003.819861}
}

@inproceedings{lpips,
  title={The Unreasonable Effectiveness of Deep Features as a Perceptual Metric},
  author={Zhang, Richard and Isola, Phillip and Efros, Alexei A. and Shechtman, Eli and Wang, Oliver},
  booktitle={Proceedings of the IEEE Conference on Computer Vision and Pattern Recognition ({CVPR})},
  pages={586--595},
  year={2018},
  doi={10.1109/CVPR.2018.00068}
}

@inproceedings{chi2021test,
  title={Test-time fast adaptation for dynamic scene deblurring via meta-auxiliary learning},
  author={Chi, Zhixiang and Wang, Yang and Yu, Yuanhao and Tang, Jin},
  booktitle={Proceedings of the IEEE/CVF conference on computer vision and pattern recognition},
  pages={9137--9146},
  year={2021}
}

@article{ye2026asmil,
  title={Asmil: Attention-stabilized multiple instance learning for whole slide imaging},
  author={Ye, Linfeng and Hamidi, Shayan Mohajer and Chi, Zhixiang and Li, Guang and Pilanci, Mert and Ogawa, Takahiro and Haseyama, Miki and Plataniotis, Konstantinos N},
  journal={arXiv preprint arXiv:2603.06658},
  year={2026}
}

@inproceedings{chi2025learning,
  title={Learning to adapt frozen clip for few-shot test-time domain adaptation},
  author={Chi, Zhixiang and Gu, Li and Liu, Huan and Wang, Ziqiang and Wu, Yanan and Wang, Yang and Plataniotis, Konstantinos},
  booktitle={International Conference on Learning Representations},
  volume={2025},
  pages={66359--66380},
  year={2025}
}

@inproceedings{wu2024test,
  title={Test-time domain adaptation by learning domain-aware batch normalization},
  author={Wu, Yanan and Chi, Zhixiang and Wang, Yang and Plataniotis, Konstantinos N and Feng, Songhe},
  booktitle={Proceedings of the AAAI Conference on Artificial Intelligence},
  volume={38},
  number={14},
  pages={15961--15969},
  year={2024}
}

@article{hinton2015distilling,
  title={Distilling the knowledge in a neural network},
  author={Hinton, Geoffrey and Vinyals, Oriol and Dean, Jeff},
  journal={arXiv preprint arXiv:1503.02531},
  year={2015}
}

@article{zhong2022meta,
  title={Meta-dmoe: Adapting to domain shift by meta-distillation from mixture-of-experts},
  author={Zhong, Tao and Chi, Zhixiang and Gu, Li and Wang, Yang and Yu, Yuanhao and Tang, Jin},
  journal={Advances in Neural Information Processing Systems},
  volume={35},
  pages={22243--22257},
  year={2022}
}

@inproceedings{hamidi2024train,
  title={How to train the teacher model for effective knowledge distillation},
  author={Hamidi, Shayan Mohajer and Deng, Xizhen and Tan, Renhao and Ye, Linfeng and Salamah, Ahmed Hussein},
  booktitle={European Conference on Computer Vision},
  pages={1--18},
  year={2024},
  organization={Springer}
}

@inproceedings{ye2024bayes,
  title={Bayes conditional distribution estimation for knowledge distillation based on conditional mutual information},
  author={Ye, Linfeng and Mohajer Hamidi, Shayan and Tan, Renhao and Yang, En-Hui},
  booktitle={International Conference on Learning Representations},
  volume={2024},
  pages={26722--26754},
  year={2024}
}

\end{document}